\documentclass[twoside]{article}
\usepackage{aistats2020} 

\usepackage[textwidth=2.3cm,textsize=small]{todonotes}
\usepackage{enumitem}
\usepackage{times}  
\usepackage{helvet} 
\usepackage{courier}  
\usepackage[hyphens]{url}  
\usepackage{graphicx} 
\usepackage[tight,footnotesize]{subfigure}
\usepackage{xcolor}
\usepackage{amsmath}
\usepackage{amsfonts}
\usepackage{mathtools}
\usepackage{cuted}
\usepackage{multicol}
\usepackage[normalem]{ulem}
\usepackage{algorithmicx}
\usepackage[ruled,vlined,linesnumbered]{algorithm2e}
\usepackage[square,comma,sort&compress,numbers]{natbib}
\usepackage{amssymb,amsthm}
\usepackage{mathtools, nccmath}
\usepackage{booktabs}

\newcommand\independent{\protect\mathpalette{\protect\independenT}{\perp}}
\def\independenT#1#2{\mathrel{\rlap{$#1#2$}\mkern2mu{#1#2}}}

\newcommand*{\ModelA}{\mathit{{M1(Linear)}}}
\newcommand*{\ModelB}{\mathit{{M2(LSTM)}}}
\newcommand*{\ModelC}{\mathit{{M3(LSTM)}}}
\newcommand*{\ModelD}{\mathit{{M4(LSTM)}}}

\usepackage{graphicx}  
\setlist[itemize]{noitemsep, topsep=0pt}
\begin{document}



\twocolumn[
\aistatstitle{G-Net: A Deep Learning Approach to G-computation for Counterfactual Outcome Prediction Under Dynamic Treatment Regimes}
\aistatsauthor{
\vspace{-0.1cm}\\Rui Li$^{*}$$^{1}$, Zach Shahn$^{*}$$^{2,3}$, Jun Li$^{1}$,  Mingyu Lu$^{1}$,  Prithwish Chakraborty$^{2,3}$,\\ Daby Sow$^{2,3}$, Mohamed Ghalwash$^{2,3}$, Li-wei Lehman$^{1}$ \\ 
$^{1}$Massachusetts Institute of Technology\\ $^{2}$IBM Research,
$^{3}$MIT-IBM Watson AI Lab
\vspace{0.3cm}} ]


\begin{abstract}
Counterfactual prediction is a fundamental task in decision-making. G-computation is a method for estimating expected counterfactual outcomes under dynamic time-varying treatment strategies. Existing G-computation implementations have mostly employed classical regression models with limited capacity to capture complex temporal and nonlinear dependence structures. This paper introduces G-Net, a novel sequential deep learning framework for G-computation that can handle complex time series data while imposing minimal modeling assumptions and provide estimates of individual or population-level time-varying treatment effects. We evaluate alternative G-Net implementations using realistically complex temporal simulated data obtained from CVSim, a mechanistic model of the cardiovascular system. 
\end{abstract}




\section{Introduction}
Counterfactual prediction is a fundamental task in decision-making.
It entails the estimation of expected future 
trajectories of variables of interest under alternative courses of action (or {\em treatment strategies})
given observed history.
Treatment strategies of interest are usually {\em time varying} (meaning they comprise decisions at multiple time points) and {\em dynamic} (meaning the treatment decision at each time point is a function of  history up to that time point). 

As an example, consider the problem of fluid administration in Intensive Care Units (ICUs)~\cite{finfer2018intravenous}. It is frequently necessary for physicians to adopt strategies that administer large volumes of fluids to increase blood pressure and promote blood perfusion through organs in septic patients. However, such strategies can lead to fluid overload, which can have serious adverse downstream effects such as pulmonary edema. Fluid administration strategies are time varying and dynamic because at each time point physicians decide the volume of fluid to administer based on observed patient history (e.g. blood pressure and volume of fluid already administered) up to that time point. To aid in the choice between alternative dynamic fluid administration strategies, it would be desirable to obtain counterfactual predictions of a patient's probability of developing fluid overload (and other outcomes of interest) were they to follow each alternative strategy going forward given their observed covariate history up to the current time. 

Counterfactual prediction is an inherently {\em causal} task in that it must account for the causal effects of following different treatment strategies. When treatment strategies of interest are time-varying, so-called ``g-methods" \cite{robins2019,robins2009} are required to estimate their effects. G-methods include g-computation \cite{robins1986,robins1987,taubman2009}, structural nested models \cite{robins1994,steelandt2014}, and marginal structural models \cite{msm2000,msm2008}. Of these methods, g-computation is best suited for estimating effects of general dynamic treatment strategies conditional on high dimensional patient histories~\cite{robins1986new}. 
%
%
%
%
%

G-computation works by estimating the conditional distribution of relevant covariates given covariate and treatment history at each time point, then producing Monte Carlo estimates of counterfactual outcomes by simulating forward patient trajectories under treatment strategies of interest. Regression model(s) for the covariates and outcomes at each time point conditional on observed history are a critical component of this method. While any regression models could in theory be input to the G-computation algorithm, most existing G-computation implementations have employed simple regression models with limited capacity to capture complex temporal and nonlinear dependence structures. 
In recent years, sequential deep learning methods such as 
Recurrent Neural Networks (RNNs)
have achieved state of the art performance in predictive modeling of complex time series data while imposing minimal modeling assumptions.
In this paper, we propose G-Net, a sequential deep learning framework for G-computation. G-Net admits the use of recurrent 
networks such as LSTMs to model covariates in a manner suitable for G-computation. 
%
The G-Net framework is flexible and allows for various configurations
depending on the problem at hand. 
To the best of our knowledge, this is the first work to investigate a RNN based approach to G-computation.

 Unfortunately, it is impossible to reliably evaluate counterfactual predictions on real data, since only the outcomes corresponding to treatment strategies that were actually followed can be observed. Consequently, to explore and evaluate various implementations of G-Net, we used simulated data in which counterfactual ground truth can be known. We used CVSim \cite{heldt2010cvsim}, a well established mechanistic model of the cardiovascular system, to estimate counterfactual simulated patient trajectories under various fluid and vasopressor administration strategies. These experiments provide a template for causal model evaluation using complex and physiologically realistic simulated longitudinal data. 

\section{Related Work}
Several recent works have proposed a deep learning framework for counterfactual prediction from observational data, including \cite{Atan2018, Alaa2017,Yoon2018}.  However, these works have mostly focused on learning point exposure as opposed to time-varying treatment effects, which are the focus of this paper.

G-computation for estimating time-varying treatment effects was first proposed by Robins~\cite{robins1986}. Illustrative applications of the general approach are provided in~\cite{taubman2009,young2011}, and summaries of g-computation (and other ``g-methods'' for estimating time-varying treatment effects) can be found in~\cite{robins2019,robins2009}. The g-computation algorithm takes arbitrary regression models as inputs. While most applications (e.g.~\cite{taubman2009,young2011}) have thus far employed classical generalized linear models, there is no conceptual barrier to using more complex machine learning regression models. 
RNNs, and in particular LSTMs, have achieved state of the art performance on a wide variety of time series regression tasks, including healthcare related tasks \cite{tomavsevNature2019,Caojamia2018,ChoiNIPS2016}. However, despite the popularity and success of RNNs for time series regression, we have not seen in the literature any ``deep" implementation of g-computation.

Recently, \citet{Lim2018} plugged RNN regression models into history adjusted marginal structural models (MSM)~\cite{ha_msm} to make counterfactual predictions. However, these MSMs can only make counterfactual predictions under $static$ time-varying treatment strategies that do not depend on recent covariate history. For example, a history adjusted MSM \citep{Lim2018} could estimate the probability of fluid overload given patient history under the (static) treatment strategy ``\textit{give 1 liter fluid each hour for the next 3 hours}'', but it could not estimate the probability of fluid overload given patient history under the (dynamic) treatment strategy ``\textit{each hour for the next 3 hours, if blood pressure is less than 65 $then$ give 1 liter fluids, otherwise give 0 liters}''. History adjusted MSMs cannot estimate effects of time-varying treatment strategies that respond to changes in the patient's health history, but g-computation can. Further, g-computation is able to straightforwardly estimate the $distribution$ of a counterfactual outcome under a time-varying treatment strategy. This is not straightforward to do with history adjusted MSMs.

\citet{Schulam2017} propose Counterfactual Gaussian Processes, an implementation of continuous time g-computation, and like us apply their method to ICU data. They only consider static time-varying treatment strategies, though it appears that their method might straightforwardly be extended to handle dynamic strategies as well. 
An advantage of Gaussian processes is interpretable uncertainty quantification. However, GPs are intractable for large datasets since they have time complexity of $O(N^{3})$, where N is the number of observations \cite{Titsias2009}. Sparse GPs, which introduce $M$ inducing points, have at least $O(M^{2}N)$ \cite{Titsias2009} time complexity. RNNs are more scalable. Recurrent dropout can also be employed in RNN based implementations to produce uncertainty estimates that approximate posterior distributions from a Gaussian process \cite{gal2}.

\section{G-computation for counterfactual prediction}\label{gcomp}

Our goal is to predict patient outcomes under various future treatment strategies given observed patient histories. 
\noindent Let:
\begin{itemize}
\item $t\in \{0,\ldots ,K\}$ denote time, assumed discrete, with $K$ being the end of followup;
\item $A_{t}$ denote the observed treatment action at time $t$;
\item $Y_t$ denote the observed value of the outcome at time $t$ 
\item $L_{t}$ denote a vector of covariates at time $t$ that may influence treatment decisions or be associated with the outcome; 
\item $\bar{X}_{t}$ denote the history $X_{0},\ldots ,X_{t}$ and $\underline{X}_{t}$
denote the future $X_{t},\ldots ,X_{K}$ for arbitrary time varying variable~$X$.
\end{itemize}

At each time point, we assume the causal ordering $(L_t,A_t,Y_t)$. Let $H_t\equiv (\bar{L}_t,\bar{A}_{t-1})$ denote patient history preceding treatment at time $t$. A dynamic treatment strategy $g$ is a collection of functions $\{g_0,\ldots,g_K\}$, one per time point, such that $g_t$ maps $H_t$ onto a treatment action at time $t$. A simple dynamic strategy for fluid volume might be $g_t(H_t)=.5\times\mathbf{1}\{bp_t<65\}$, i.e. give .5 liters of fluid if mean arterial blood pressure is less than 65 at time~$t$. 

Let $Y_t(g)$ denote the counterfactual outcome that would be observed at time $t$ had, possibly contrary to fact, treatment strategy $g$ been followed from baseline \cite{robins1986}. Further, let $Y_t(\bar{A}_{m-1},\underline{g}_m)$ with $t\geq m$ denote the counterfactual outcome that would be observed had the patient received their observed treatments $\bar{A}_{m-1}$ through time $m-1$ then followed strategy $g$ from time $m$ onward. 


In counterfactual point prediction, our goal is to estimate expected counterfactual patient outcome trajectories 
\begin{align}\label{target}
\{E[Y_t(\bar{A}_{m-1},\underline{g}_m)|H_m], t\geq m\}
\end{align}
given observed patient history through time $m$ for any $m$ and any specified treatment strategy $g$. We might also be interested in estimating the counterfactual outcome distributions at future time points
\begin{align}\label{target2}
\{p(Y_t(\bar{A}_{m-1},\underline{g}_m)|H_m), t\geq m\}.
\end{align}

If we do not condition on anything in $H_m$, then (\ref{target}) is an expectation (and (\ref{target2}) a distribution) over the full population. If we condition on a small subset of variables contained in patient history, then (\ref{target}) is an expectation (and (\ref{target2}) a distribution) over a sub-population. If we condition on all elements of a patient history, then (\ref{target}) is still technically only an expectation (and (\ref{target2}) a distribution) over a hypothetical sub-population with the exact patient history conditioned on, but in this case (\ref{target}) and (\ref{target2}) practically amount to what is usually meant by personalized prediction. 

Under the below standard assumptions, we can estimate (\ref{target}) and (\ref{target2}) through g-computation \cite{robins1986}.
\begin{enumerate}[labelsep=2pt, leftmargin=10pt, topsep=0pt,itemsep=0ex,partopsep=1ex,parsep=1ex]
    \item \textbf{Consistency:} $\bar{Y}_K(\bar{A}_K) = \bar{Y}_K$
    \item \textbf{Sequential Exchangeability}: $\underline{Y}_t \independent A_t|H_t\; \; \forall t$
    \item \textbf{Positivity:} $P(A_t=g_t(H_t))>0\; \forall\{H_t: P(H_t)>0\}$
\end{enumerate}








Assumption 1 states that the observed outcome is equal to the counterfactual outcome corresponding to the observed treatment. Assumption 2 states that there is no unobserved confounding of treatment at any time and any future outcome. Assumption 2 would hold, for example, under the conditions depicted in Figure 1. Positivity states that the counterfactual treatment strategy of interest has some non-zero probability of actually being followed. Under the assumption that we specify certain predictive models correctly such that their predictions extrapolate to parts of a joint distribution that they were not trained on, positivity is not strictly necessary. 
\begin{figure}
    \centering
    \includegraphics[scale=0.10]{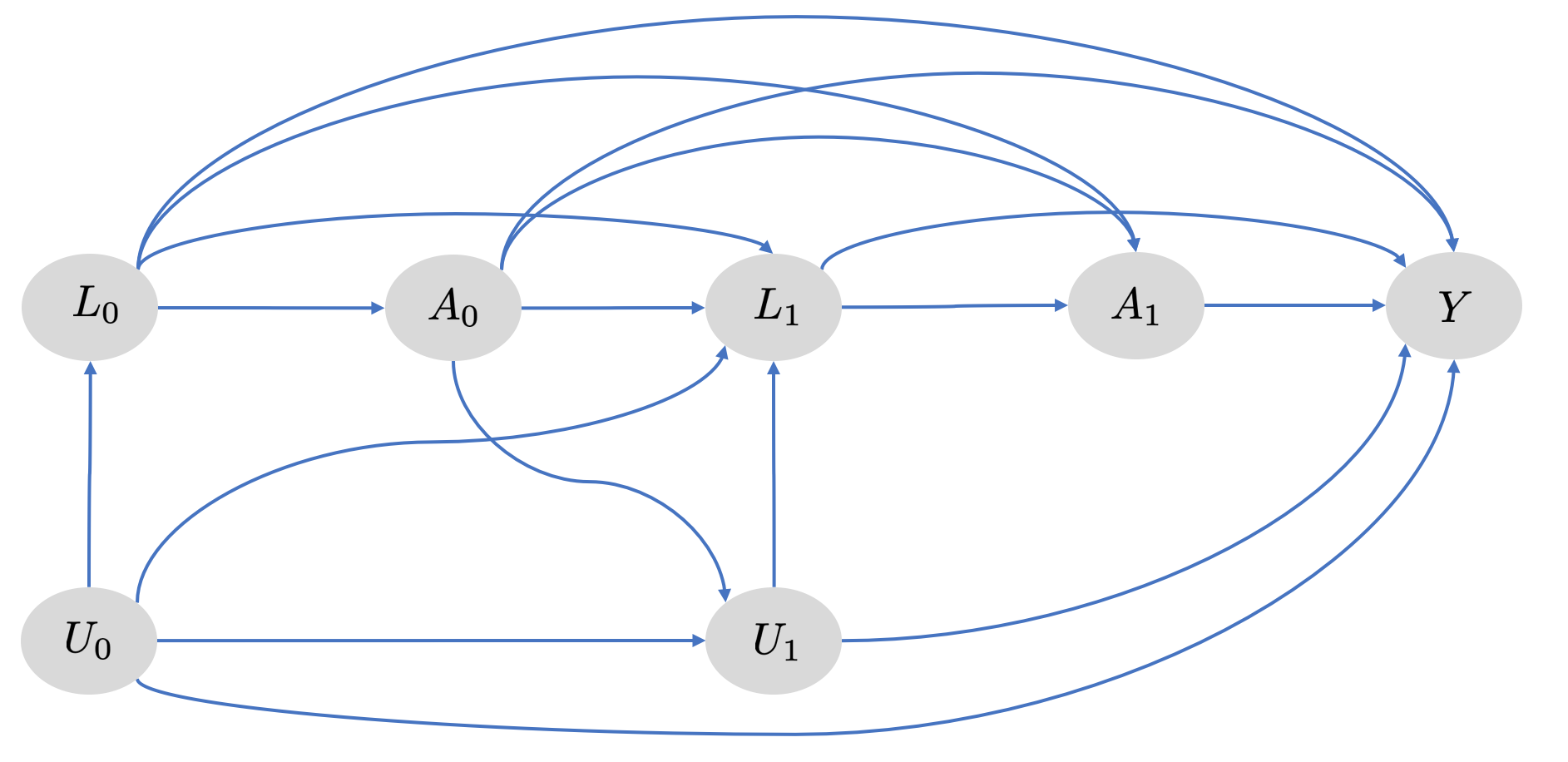}
    \caption{\small{A causal DAG representing a data generating process in which Assumption 2 (sequential exchangeability) holds. This DAG represents a simple two time step process where the outcome is measured only after the final time step. However, its salient property is that all variables influencing treatment (i.e. with arrows directly into treatment) and associated with future outcomes are measured.}}
    \label{fig:my_label}
\end{figure}

Under assumptions 1-3, for $t=m$ we have simply that 
{\small
\begin{align}
    p(Y_m(\bar{A}_{m-1},g_m)|H_m) = p(Y_m|H_m,A_m=g_m(H_m)),
\end{align}
}%
i.e. the conditional distribution of the counterfactual is simply the conditional distribution of the observed outcome given patient history and given that treatment follows the strategy of interest. For $t>m$, things are slightly more complex because we need to adjust for time-varying confounding. With $X_{i:j} = X_i,\ldots,X_j$ for any random variable X, under Assumptions 1-3 the g-formula yields

{\small
\begin{align}\label{g_dist}
& p(Y_t(\bar{A}_{m-1},\underline{g}_m)=y|H_m) \nonumber \\
& = \int_{l_{m+1:t}} p(Y_t=y|H_m,L_{m+1:t}=l_{m+1:t}, A_{m:t}=g(H_{m:t})) \nonumber \\
& \quad\quad\quad\quad \times \prod_{j=m+1}^t p(L_j = l_j|H_m,L_{m+1:j-1}=l_{m+1:j-1},\nonumber\\
& \quad\quad\quad\quad\quad\quad\quad\quad A_{m,j-1}=g(H_m,l_{m+1:j-1})).
\end{align}
}%


It is not generally possible to compute this integral in closed form, but it could be approximated through Monte-Carlo simulation. We repeat the recursive process shown in Algorithm \ref{alg:gcomp} $M$ times. (Here the outcome $Y_t$ is without loss of generality deemed to be a variable in the vector $L_{t+1}$.)
\begin{figure*}[!bthp]
    \centering
    \includegraphics[scale=.14]{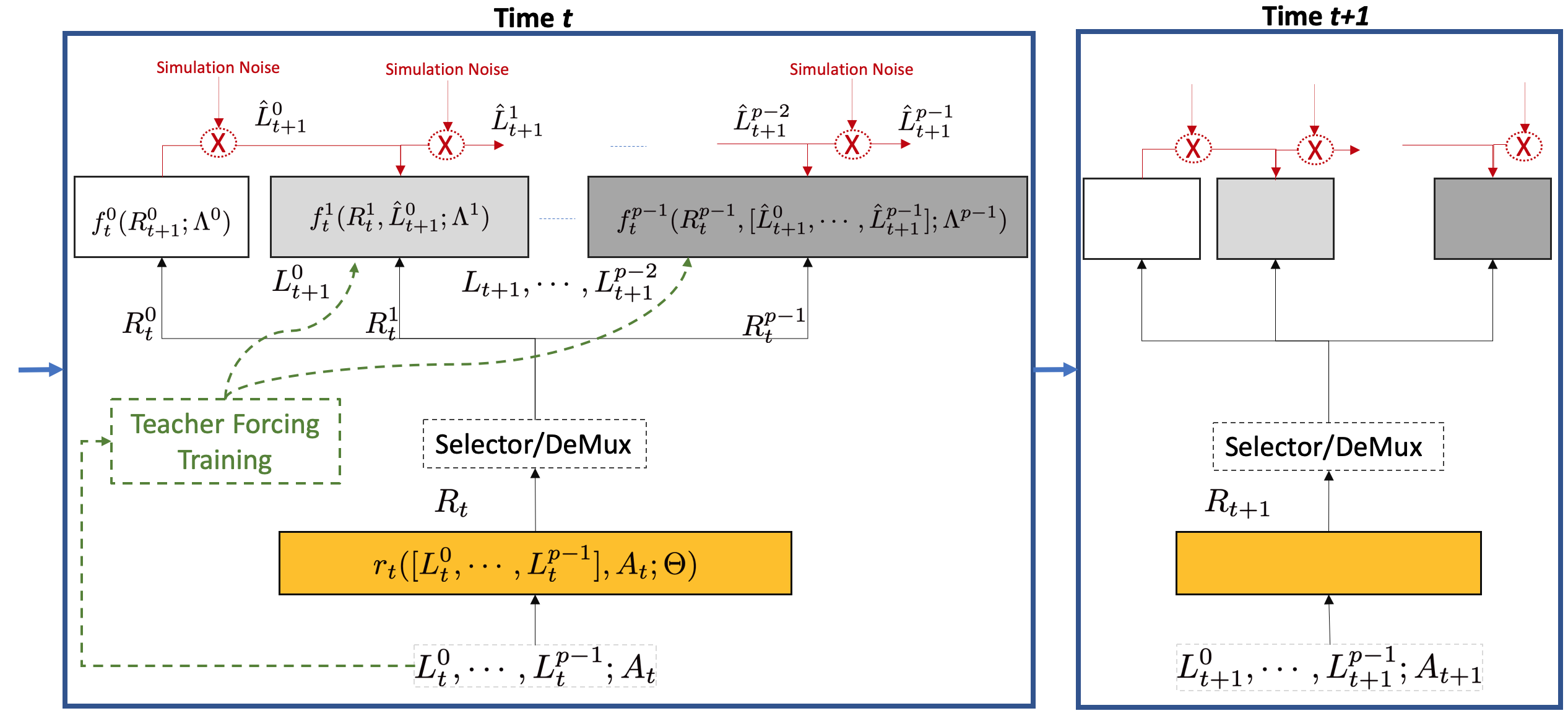}
    \caption{\small{The G-Net: A flexible sequential deep learning framework for g-computation.}}
    \label{fig:gen_framework}
\end{figure*}

At the end of this process, we have $M$ simulated draws of the counterfactual outcome for each time $t=\{m,\ldots,K\}$. For each $t$, the empirical distribution of these draws constitutes a Monte-Carlo approximation of the counterfactual outcome distribution (\ref{target2}). The sample averages of the draws at each time $t$ are an estimate of the conditional expectations (\ref{target}) and can serve as point predictions for $Y_t(\bar{A}_{m-1},\underline{g}_m)$ in a patient with history $H_m$.

Key to the g-computation algorithm is the ability to simulate from joint conditional distributions $p(L_{t} |\bar{L}_{t-1},\bar{A}_{t-1})$ of the covariates given patient history at time $t$. Of course, in practice we do not have prior knowledge of these conditional distributions and need to estimate them from data. Most implementations use generalized linear regression models to estimate the conditional distributions of the covariates. Often, these models do not capture temporal dependencies present in the patient data. We propose the G-Net for this task.

\begin{algorithm}[h]
\DontPrintSemicolon
\SetKwInOut{Input}{Input}
\SetKwInOut{Output}{Output}
\caption{G-Computation (One simulation)}\label{alg:gcomp}
Set $a_m^*=g_m(H_m)$\;
Simulate and record $y_m^*$ from $p(Y_m|H_m,A_{m}=a_{m}^*)$\;
Simulate $l_{m+1}^*$ from $p(L_{m+1} |H_m,A_{m}=a_{m}^*)$\;
Set $a_{m+1}^*=g_m(H_m,l_{m+1}^*,a_{m}^*)$\;
Simulate and record $y_{m+1}^*$ from $p(Y_{m+1}|H_m,L_{m+1}=l_{m+1}^*,A_m=a_m^*,A_{m+1}=a_{m+1}^*)$\;
Simulate $l_{m+2}^*$ from $p(L_{m+2}|H_{m},L_{m+1}=l_{m+1}^*,A_{m}=a_{m}^*,A_{m+1}=a_{m+1}^*)$\;
Continue simulations through time $K$
\end{algorithm}

\section{G-Net Design}

The proposed G-Net framework depicted in Figure~\ref{fig:gen_framework} enables the use of sequential deep learning models to estimate conditional distributions 
$p(\bar{L}_t|\bar{L}_{t-1},\bar{A}_{t-1})$ 
of covariates given history at each time point and perform the G-computation algorithm described in Algorithm~\ref{alg:gcomp} to simulate covariates under various treatment strategies. In this setting, without loss of generality, we set $Y_t$ as one of the co-variates in $L$ for notational simplicity.   

Let $L_t^0,\ldots,L_t^{p-1}$ denote $p$ components of the vector $L_t$, where each component $L_t^j$ could be multivariate.
%
%
%
We impose an arbitrary ordering \sloppy $\medmath{L_1^0,L_1^1,L_1^2,\ldots,L_1^{p-1},A_1,\ldots,L_K^0, L_K^1,L_K^2,\ldots,L_K^{p-1},A_k}$ and estimate the conditional distributions of each $L_t^j$ given all variables preceding it in this ordering. At simulation time, we exploit the basic probability identity

{
  $ \displaystyle
    \begin{aligned} 
\medmath{p(L_{t} |\bar{L}_{t-1},\bar{A}_{t-1})} &= 
\medmath{p(L_t^0 |\bar{L}_{t-1},\bar{A}_{t-1})} \times 
\medmath{p(L_{t}^1 |L_t^0,\bar{L}_{t-1},\bar{A}_{t-1})}\\
& \times \dots \times 
\medmath{p(L_{t}^{p-1} |L_t^0,\ldots,L_t^{p-2},\bar{L}_{t-1},\bar{A}_{t-1})}
    \end{aligned}
  $ 
\par}%

to simulate from $p(L_{t} |\bar{L}_{t-1},\bar{A}_{t-1})$ by sequentially simulating each $L_t^j$ from $p(L_t^j|L_t^0,\ldots,L_t^{j-1},\bar{L}_{t-1},\bar{A}_{t-1})$.
There are at least two reasons to allow for subdivision of the covariates. First, if covariates are of different types (e.g. continuous, categorical, count, etc.), it is difficult to simultaneously simulate from their joint distribution. Second, customizing models for each covariate component can potentially lead to better performance. 
Figure~\ref{fig:gen_framework} illustrates this decomposition where at each time point the components are depicted via ordered (grey shaded) boxes that are responsible for the estimation of the various terms needed to compute the conditional distributions.
One could set $p=1$ and model all covariates simultaneously or at the other extreme set $p$ to be the total number of variables.

The sequential model used in G-Net provides us with estimates of the conditional expectations \sloppy $E[L_t^j|\bar{L}_{t-1},L_t^0,\ldots,L_t^{j-1},\bar{A}_{t-1}]$ for all $t$ and $j$. To simulate from $p(L_t^j|\bar{L}_{t-1},L_t^0,\ldots,L_t^{j-1},\bar{A}_{t-1})$, we proceed as follows. If $L_t^j$ is multinomial, its conditional expectation defines its conditional density. If $L_t^j$ has a continuous density, there are various approaches we might take to simulate from its conditional distribution. Without making parametric assumptions, we could simulate from

\begin{align}
\label{sim}
L_{t}^j|L_t^0,\ldots,L_t^{j-1},\bar{L}_{t-1},\bar{A}_{t-1}\sim \qquad\qquad\qquad \nonumber\\ 
\qquad\qquad \hat{E}[L_{t}^j |L_t^0,\ldots,L_t^{j-1},\bar{L}_{t-1},\bar{A}_{t-1}] + \epsilon_t^j
\end{align}
where $\epsilon_t^j$ is a draw from the empirical distribution of the residuals $L_t^j - \hat{L}_t^j$ in a holdout set not used to fit the model parameters used to generate $\hat{L}_t^j$ as an estimate of $E[L_{t}^j |L_t^0,\ldots,L_t^{j-1},\bar{L}_{t-1},\bar{A}_{t-1}]$. This method makes the simplifying assumption that the covariate error distribution does not depend on patient history. This is the approach we take in the experiments in this paper, and is depicted in the simulation noise nodes at the top of Figure~\ref{fig:gen_framework}.
Alternatively, we might specify a parametric distribution for $L_t^j - E[L_{t}^j |L_t^0,\ldots,L_t^{j-1},\bar{L}_{t-1},\bar{A}_{t-1}]$, e.g. a Gaussian, and directly estimate its parameters by maximum likelihood. 


To obtain the estimates for the conditional distribution of covariates, 
G-Net admits sequential modeling of patients' data using the group
decomposition and arbitrary ordering outlined before.
As shown in the yellow boxes in Figure~\ref{fig:gen_framework}, at each time $t$, a representation $R_t$ of the patient history can be computed as
\begin{equation}
    R_t = r_t(\bar{L}_t,\bar{A}_t;\Theta)
\end{equation}
where $\Theta$ represents model parameters learned during training. 
In its simplest form, $r_t$ may just be an identity function passing the covariates. 
In other configurations (e.g. using the selector in Figure~\ref{fig:gen_framework}),
$r_t$ can be used to provide abstractions of histories using sequential learning architectures
such as RNN. 
This formulation of $r_t$ allows for a great deal of flexibility in how information is shared across variables and time. 

Estimates from each of the $p$ covariate groups can then be obtained, as shown in Figure~\ref{fig:gen_framework},
by successive estimation of conditional expectations of covariates.  Specifically, the conditional expectation of each $L_{t+1}^j, 0\leq j < p$ given the representation of patient history $R_t$ and the other variables from time $t+1$ that precede it in the arbitrary predefined ordering is estimated by the functions $f_{t}^j$. The complete sequence can be given
as follows:
\begin{align}
 &L_{t+1}^0 = f_{t}^0(R_t;\Lambda_{0}) \nonumber\\
 &L_{t+1}^1 = f_{t}^1(R_t, L_{t+1}^0;\Lambda_1) \nonumber \\
 &\cdots \nonumber \\
 &L_{t+1}^j = f_{t}^j(R_t,[L_{t+1}^1,\ldots,L_{t+1}^{j-1}];\Lambda_{j}) \nonumber \\
 &\cdots
\label{eq:r_t_gen}
\end{align}
where, each of the $f_{t}^j$ represents the specialized estimation function for group $j$ and $\Lambda_j$ are the learnable parameters for the same. 
Depending on modeling choices, $f_t^j$ can be sequential models specialized for group $j$ or models focusing only on the time step $t$ (e.g. linear models).\\
\\
Given G-Net parameters, the distribution of the Monte Carlo simulations produced by G-computation Algorithm 1 constitute an estimate of uncertainty about a counterfactual prediction. But this estimate ignores uncertainty about the G-Net parameter estimates themselves. One way to incorporate such model uncertainty would be to fit a Bayesian model and, before each Monte Carlo trajectory simulation in G-computation, draw new network parameters from their posterior distribution. These Monte Carlo draws would be from the posterior predictive distribution of the counterfactual outcome. Bayesian deep learning can be prohibitively computationally intensive, but can be approximated through dropout \cite{gal2}. First, we fit the RNN component of the G-Net using recurrent dropout, with dropout masks at each layer held constant across time as described in \cite{gal2}. Let $M$ denote a dropout mask constant across time for each layer of the RNN, $\hat{\Theta}(M)$ denote estimated RNN parameters with mask $M$ applied, and $p_M$ denote the distribution from which $M$ was sampled during training. Then, during g-computation, we add \textit{step 0} to the beginning of Algorithm 1 before each simulation:
\begin{align*}
    &\textbf{Step 0 (dropout)}: \text{Sample } D*\sim p_D 
\end{align*}
and compute all simulations plugging $\hat{E}_{\Theta(D)}[L_{t}^j |L_t^1,\ldots,L_t^{j-1},\bar{L}_{t-1},\bar{A}_{t-1}]$ into (\ref{sim}). Then draws of $\hat{\Theta}(D)$ are from an approximation to the posterior distribution of $\Theta$ under a particular Gaussian Process prior. Therefore, the Monte Carlo simulations obtained from g-computation incorporating dropout (i.e. adding step 0 as above) approximate draws from the posterior predictive distribution of the counterfactual outcome.

%
The parameters, $\Theta$ and $\Lambda$, are learned by optimizing a loss function forcing the G-Net to accurately estimate covariates $L_t$ at each time point $t$ using standard gradient descent techniques. It is to be noted that it is necessary to use teacher-forcing (using observed values of $L_{t+t}^{j}$ as arguments to $f_t$ in equation~\label{eq:r_t_gen}) to obtain unbiased estimates of $f_t$, as shown in Figure \ref{fig:gen_framework}. 

\section{Simulation Experiments Using CVSim}

To evaluate counterfactual predictions, it is necessary to use simulated data in which counterfactual ground truth for outcomes under alternative treatment strategies is known. To this end, we performed experiments on data generated by CVSim \cite{CVSim2010}, a program that simulates the dynamics of the human cardiovascular system. 

\begin{figure}
\includegraphics[width=0.45\textwidth,height=0.26\textheight]{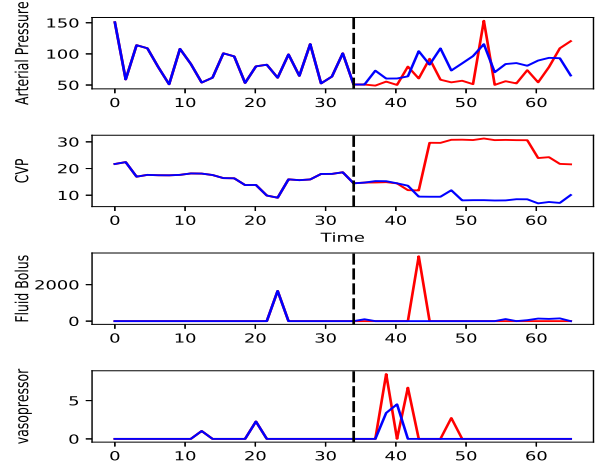}
\caption{\small{Covariates trajectories for the same patient (i.e. the same random seed) under two different treatments $g_{o}$ (blue) and $g_{c}$ (red). After the treatment strategies diverge at middle of trajectory, $t = 34$ (black dashed line), $g_{c}$ delivers a large fluid dose while $g_{o}$ does not, temporarily pushing AP and CVP higher under the $g_{c}$ regime than the $g_{o}$ regime.} }
\label{fig:scvsim}
\end{figure}

\textbf{Data Generation:} We generated an `observational' dataset $D_{o}$ under treatment regime $g_{o}$ and two `counterfactual' datasets $D_{c1}$ and $D_{c2}$ under treatment regimes $g_{c1}$ and $g_{c2}$. The data generating processes producing $D_{o}$ and $D_{cj}$ were the same except for the treatment assignment rules. For each $j$, $g_{cj}$ was identical to $g_{o}$ for the first $m-1$ simulation time steps before diverging to a different treatment rule for time steps $m$ to $K$ as illustrated in Figure~\ref{fig:scvsim}. 

A CVSim 6-compartment circulatory model takes as inputs 28 variables that together govern a hemodynamic system. It then deterministically simulates forward in time a set of 25 output variables according to a collection of differential equations (parameterized by the input variables) modeling hemodynamics.
Important variables in CVSim  include arterial pressure (AP), central venous pressure (CVP), total blood volume (TBV), and total peripheral resistance (TPR). In real patients, physicians observe AP and CVP and seek to keep them above a clinically safe threshold. They do this by intervening on TBV (through fluid administration) and TPR (through vasopressors).

We defined simulated treatment interventions that were designed to mimic the impact of fluids and vasopressors. These simulated interventions alter the natural course of the simulation by increasing either TBV (in the case of the simulated fluids intervention) or TPR (in the case of the simulated vasopressor intervention). We generated patients by randomly initiating baseline inputs (which we hid from our G-Nets to make this a stochastic modeling problem) within plausible physiologic ranges, then using CVSim to simulate covariates forward, intervening according to the relevant treatment strategy at each timestep. Full details of the simulation process can be found in the Appendix.

Under (stochastic) observational treatment strategy $g_o$, the probability of receiving a non-zero vasopressor or fluid dose at a given time increases as MAP and CVP decrease according to a logistic regression function. Given that a dose is non-zero, the exact amount is drawn from a normal distribution with mean inversely proportional to MAP and CVP. Since all drivers of treatment under $g_o$ are observed in our data, the sequential exchangeability assumption holds and g-computation may be validly applied. 

$g_{c1}$ is similar to $g_o$, except it is a deterministic treatment strategy and the functions linking treatment and dose to covariates have different coefficents. Under $g_{c2}$, treatment is always withheld. Again, details are in the Appendix.

\textbf{Experiment Setup:} We set ourselves the task of training a G-Net on $D_{o}$ and using it to predict the trajectories of patients in $D_{cj}$ for time steps $m$ to $K$ for each $j$.  
This setup is designed to evaluate the performance of a G-Net in a situation in which we observe data from past patients ($D_{o}$) who received usual care ($g_{o}$) for K timesteps and would like to predict how a new patient who has been observed for $m$ timesteps would fare were they to follow a different treatment strategy of interest ($g_{cj}$) for timesteps $m$ to $K$. This is a standard use case for counterfactual prediction. $D_{cj}$ provides ground truth for a collection of patients whose trajectories follow just the path we are interested in predicting. Thus, by aggregating predictive performance metrics across simulated patients in $D_{cj}$ we generate measures of the population level performance of our G-Net at the counterfactual prediction task for which it was intended.

\begin{table}[h!]
\caption{\small{Experimental Model Setup Grid: Each cell summarizes the instantiations: $\ModelA$, $\ModelB$, $\ModelC$, and $\ModelD$), of G-Net used in our experiments.} \label{tab:model_grid}} 

\begin{center}
\small
    \begin{tabular}{cll}  
    \toprule
      {} & \textbf{With pass thru} $r_t$ & \textbf{With sequential $r_t$} \\
     \midrule
                    &  {$\underline{\ModelA}$}  & {$\underline{\ModelB}$}\\
        {$f_i$:LR}  &  $\bullet$ $r_t$: identity &  $\bullet$ $r_t$: LSTM \\
                    &  $\bullet$ $p = 2$ &  $\bullet$ $p = 2$\\
                    &  $\bullet$ $(f_0,f_1)$: linear layers. & $\bullet$ $(f_0,f_1)$: linear layers. \\ 
      \midrule
                    &  {$\underline{\ModelC}$}  & {$\underline{\ModelD}$}\\
       {$f_i$:RNN}  & $\bullet$ $r_t$: identity &  $\bullet$ $r_t$: LSTM \\
                    &  $\bullet$ $p = 2$ &  $\bullet$ $p = 2$\\
                    &  $\bullet$ $(f_0,f_1)$: LSTMs. &  $\bullet$ $(f_0,f_1)$: LSTMs \\             
     \bottomrule
    \end{tabular}
\end{center}
\end{table}
As shown in Table \ref{tab:model_grid}, we explore two specific criteria: (a) using sequential vs.\ Identity functions for $r_t$ and 
(b) using sequential models of entire patient history vs.\ linear models focusing only on current time point for $f_t$. This provides us $4$ different implementations of G-Net.
The best parameters for each model found from grid search are as follows. 
For $\ModelB$ 
the hidden dimension for the representational LSTM is $30$, the hidden dimension for the categorical LSTM is $5$, the hidden dimension for the continuous LSTM is $30$, and the learning rate is $.001$. 
For $\ModelC$, 
the hidden dimension for the categorical LSTM is $10$, the hidden dimension for the continuous LSTM is $75$, and the learning rate is $.005$. For $\ModelD$, the hidden dimension for the representation LSTM is $30$, the hidden dimension for the categorical LSTM is $5$, the hidden dimension for the continuous LSTM is $30$, and the learning rate is $.001$. 
For all models, the batch size used was 64. 
These parameters were the ones used to achieve the results presented in the experiments and results section.

\begin{figure}
\centering
\includegraphics[scale=.45]{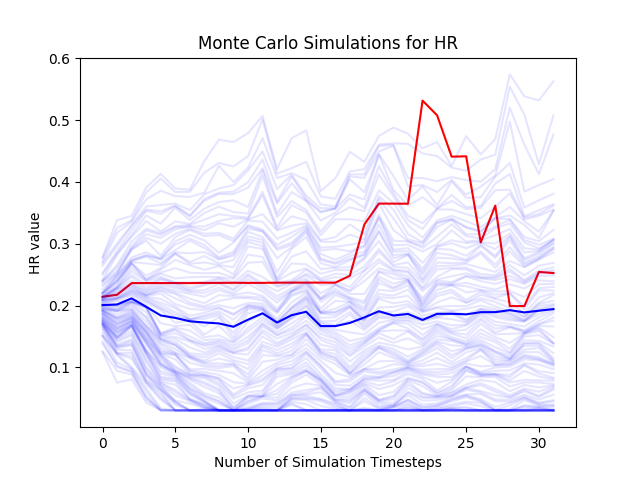}
\caption{\small{100 G-Net simulated trajectories (blue) and ground truth (red) for one patient under $g_{C1}$. } }
\label{fig:cvsim_counterfactual_example1}
\end{figure}

\begin{figure*}
\centering
\subfigure[MSE $g_{c1}$]{\includegraphics[width=0.36\linewidth]{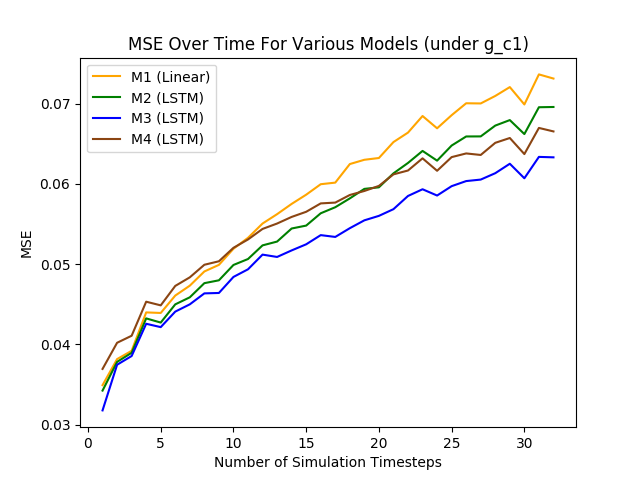}}
\subfigure[MSE $g_{c2}$]{\includegraphics[width=0.36\linewidth]{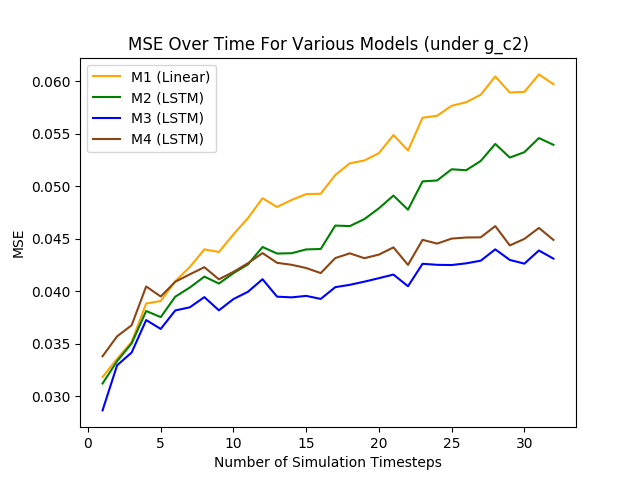}}
\subfigure[Calibration $g_{c1}$]{\includegraphics[width=0.36\linewidth]{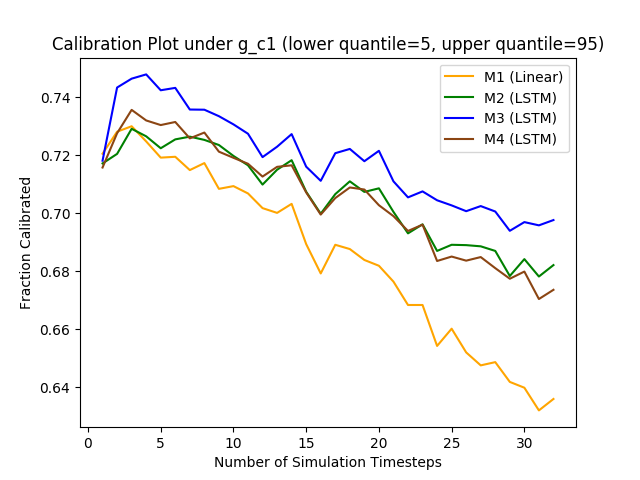}}
\subfigure[Calibration $g_{c2}$]{\includegraphics[width=0.36\linewidth]{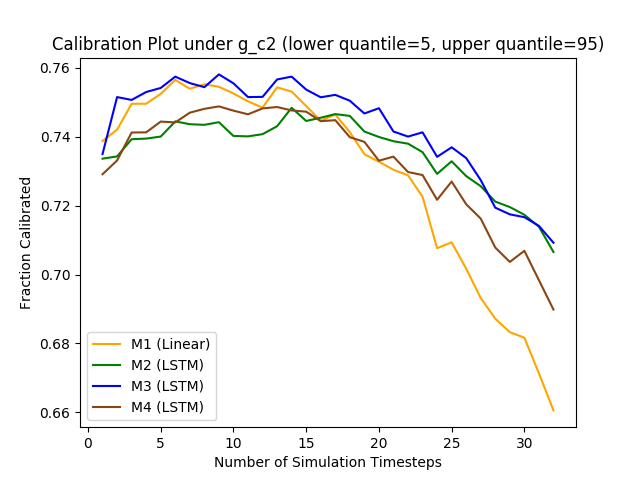}}
\caption{\small{Performance of various models: MSE and calibration over time for $g_{c1}$ and $g_{c2}$}}
\label{fig:MSE}
\end{figure*}


\textbf{Evaluation:} We evaluate the accuracy (Mean Squared Error - MSE) and calibration of the counterfactual simulations generated by our G-Nets as follows. Say $D_{cj}$ comprises $N_{c}$ trajectories of random variable $(\bar{L}_K^{cj},\bar{A}_K^{cj})$. Given observed history $H_{mi}^{cj}=(\bar{L}_{mi}^{cj},\bar{A}_{m-1i}^{cj})$ for patient $i$ from $D_{cj}$, a G-Net $G$ fit to $D_{o}$ produces $M$ (in our experiments, 100) simulations of the counterfactual covariate trajectory $\{\tilde{L}_{ti}^{cj}(H_{mi}^{cj},G,k):t\in m:K; k\in1:M\}$. These simulated trajectories are the light blue lines in Figure \ref{fig:cvsim_counterfactual_example1}.

The G-Net's point prediction of $L_{ti}$ is its estimate of $E[L_t(g_{cj})|H_m=H_{mi}]$, i.e. the average of the $M$ simulations $\hat{L}_{ti}(G)\equiv\frac{1}{M}\sum_{k=1}^{M}\tilde{L}_t^{c}(H_{mi}^{c},G,k)$. This is the dark blue line in Figure \ref{fig:cvsim_counterfactual_example1}.

If $L_t$ has dimension $d$, we compute the MSE of counterfactual predictions by a G-Net G in the dataset $D_{c}$ as $\frac{1}{N_c(K-m)d}\sum_{i=1}^{N_{c}}\sum_{t=m}^K\sum_{h=1}^d (L_{ti}^{h,CF}-\hat{L}^{h,CF}_{ti}(G))^2$.

We assess the calibration of a G-Net $G$ as follows. Given lower and upper quantiles $\alpha_{low}$ and $\alpha_{high}$, the calibration measures the frequency with which the actual counterfactual covariate $L_{ti}^{h,cj}$ is between the $\alpha_{low}$ and $\alpha_{high}$ quantiles of the $M$ simulations $\{\tilde{L}_{ti}^{h,cj}(H_{mi}^{cj},G,k):k\in1:M\}$. If this frequency is approximately $\alpha_{high}-\alpha_{low}$, then $G$ is well calibrated.

\textbf{Experiments/Results: }Using CVSim, we generated a total of 10,000 trajectories in  $D_{o}$ ($N_{o}=10,000$), of which 80\% were used for training, and the remaining 20\% for validation.  For testing, we generated 500 observations in the $D_{cj}$ datasets ($N_{c}=500$).  We included a total of 18 output variables (including all variables influencing treatment assignment under $g_o$) from CVSim to construct $D_{o}$ and $D_{cj}$; each trajectory is of length 64 time steps (d=18, K=64). In  each $D_{cj}$, the switching time point $m$ from $g_{o}$ to $g_{c}$ is fixed at 34 for all trajectories ($m=34$).   


\begin{figure*}
\centering
\includegraphics[width=0.34\linewidth]{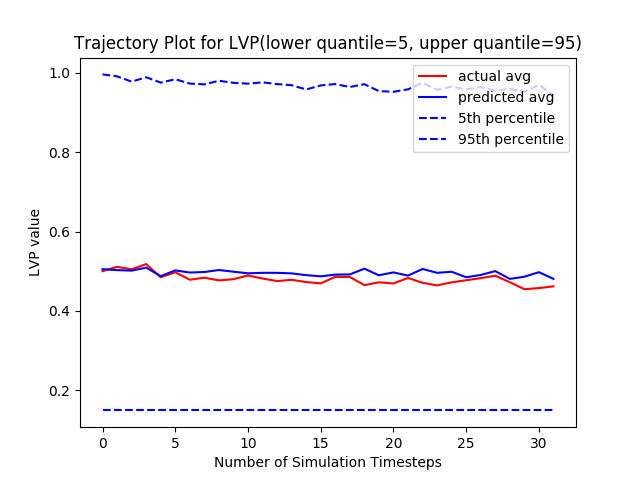}
\includegraphics[width=0.34\linewidth]{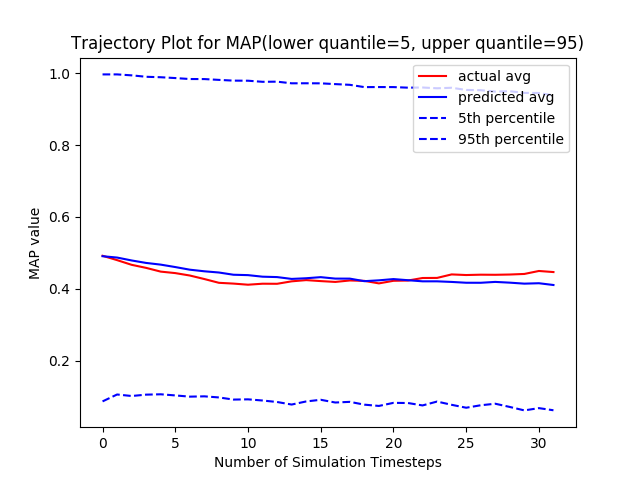}
\caption{\small{Estimated (G-Net $M3$ in Table 1) and actual population average trajectories under $g_{c1}$ for select variables.}}
\label{fig:est_vs_actual_traj}
\end{figure*}

\begin{figure*}
    \centering
    \includegraphics[width=0.34\linewidth]{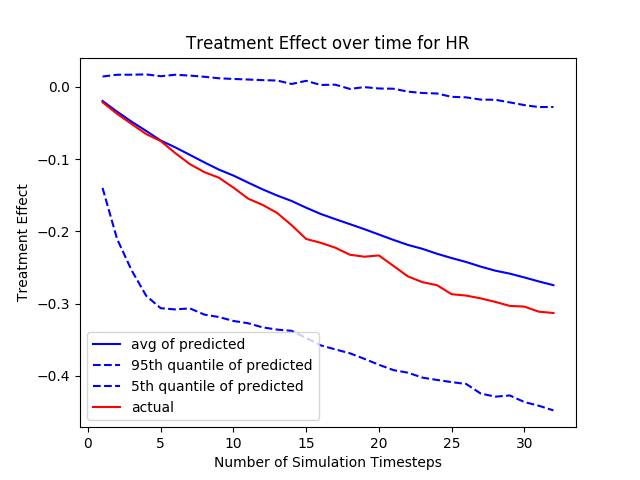}
    \includegraphics[width=0.34\linewidth]{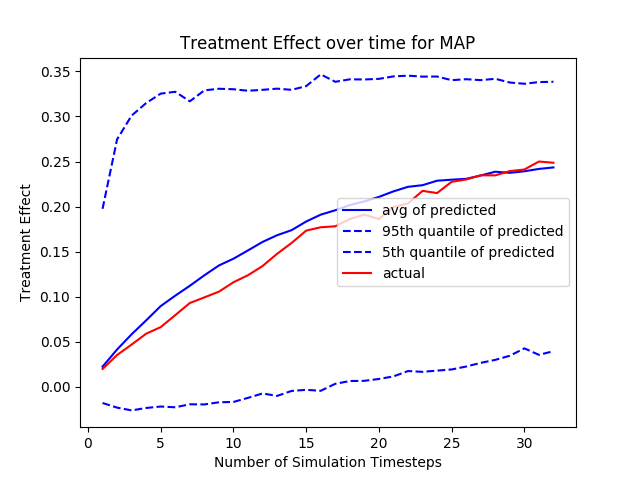}
    \caption{\small{Treatment Effect for Selected Variables}}
    \label{fig:te_var}
\end{figure*}

We fit the four models described in Table 1 to the training portion of $D_{o}$ (80\%), and used the remaining portion as validation to tune our model hyperparameters. 
Next, given observed covariate history through 34 time steps and treatment history through 33 time steps of each trajectory in each $D_{cj}$, we computed the MSE and calibration of the G-Nets' counterfactual predictions for time steps 34 to 64. 

Figure \ref{fig:MSE} (a) and (b) illustrates the performance of the various G-Net architectures in terms of MSE over time. Overall, for this experiment we found M3 (Identity function for $r_t$ and LSTM for the $f_i$ functions), performed best over both treatment strategies. Also, a comparative analysis of M1 vs M3 (both with Identity representation function for $r_t$) and  M2 vs M4 (both with LSTM representation functions for $r_t$) showed that G-Net provide better estimates of conditional distributions by admitting sequential prediction models focusing on the entire patient history compared to prediction models focusing on a single time point.

Figure 5 (c) and (d) depicts calibration for the candidate models. All of the calibration coverage rates are below the nominal levels (they should be .5), though the RNN based G-Nets again perform better than the linear model implementation. This is in part because the counterfactual predictive density estimates in these experiments do not take into account uncertainty about model parameter estimates, which could be addressed with dropout as discussed in Section 4. 



The G-Net can also be used to estimate population average counterfactual outcomes and treatment effects, quantities sometimes more relevant to policy decisions than individual level counterfactual predictions. Figure \ref{fig:est_vs_actual_traj} displays G-Net (M3 from Table 1) estimates and true values of population average trajectories for select variables under $g_{c1}$. Figure \ref{fig:te_var} shows estimates and true values of the population average treatment effect of following $g_{c1}$ as opposed to $g_{c2}$ on select variables. We see that the G-Net does a good job of estimating these population level quantities of interest.


\section{Discussion and Future Work}

In the G-Net we have introduced a novel and flexible framework for counterfactual prediction under dynamic treatment strategies through G-computation using sequential deep learning models. We have illustrated several implementations of the G-Net in a realistically complex simulation setting where we had access to ground truth counterfactual outcomes. In particular, we considered alternative approaches to representation learning that either share representations of patient history across predictive tasks or keep them separate. In the particular implementation we considered, shared representations seemed to aid simple linear classifiers but harm LSTMs.

The G-Net framework's flexibility means that there are many other alternative implementations to explore. For example, we might consider alternative architectures for representing patient history, such as attention mechanisms or memory networks.

Another direction of future work is incorporation of prior causal knowledge. For example, if it is known which variables are confounders and which merely predictive of the outcome, we might include weights in the loss function emphasizing faithful representation and prediction of confounding variables compared to non-confounders. 

\bibliography{main}
\bibliographystyle{unsrtnat}

\clearpage
\appendix
\section{Appendix}

\subsection{CVSim}
CVSim is a open-source cardiovascular simulator, aiming for education and research purpose developed by Thomas et al\cite{heldt2010cvsim}. In this work, we focus on CVSim-6C which consists of 6 components, functioning as pulmonary and systemic veins, arteries, and micro-circulations respectively. CVSim-6C is
regulated by arterial baroreflex system to simulate Sah’s lumped hemodynamic model\cite{heldt2010cvsim}. The aggregate model is capable of simulating pulsatile waveforms, cardiac output and venous return curves, and spontaneous beat-to-beat hemodynamic variability. In this work, we  modified  and  built  on  CVSim  by  adding  stochasticcomponents to it for the purposes of evaluating our coun-terfactual simulators. We call our stochastic simulation engine  S-CVSim.

\subsection{Inputs of S-CVSim}
By varying hemodynamic parameters of CVSim, it can simulate cardiovascular system under various conditions. 
We, therefore, sample values of a subset of model parameters while initiating simulation at time 0 to obtain ideal distribution of trajectories. These model parameters are listed at \textbf{Table 2}. \textbf{Note that this does not necessarily mean values of input covariates would not exceed or drop below such range at any time $t$, where $t > $ 0.}.

\begin{table}[h!]
    \caption{Names and corresponding ranges of input parameters of S-CVSim. }
    \begin{center}
    \begin{tabular}{l|l}
    \textbf{INPUT COVARIATES}  &\textbf{RANGE} \\
    \hline \\
     Total Blood Volume & 1,500 - 6,000 \\
     Nominal Heart Rate & 40 - 160 \\
     Total Peripheral Resistance & 0.1 - 1.4 \\
     Arterial Compliance & 0.4 - 1.1 \\
     Pulmonary Arterial Compliance & 0.1 - 19.9 \\
     Total Zero-pressure filling Volume & 500 - 3,500 \\
     Pulmonary Arterial Compliance & 2.0 - 3.4 \\
     Pulmonary Microcirculation Resistance & 0.4 - 1.00 \\
\end{tabular}
\end{center}

    \label{tab:my_label}
\end{table}

\subsection{Outputs of S-CVSim}

At each time $t$, CVSim-6C generates hemodynamic data including vascular resistance and varieties of type of flow, pressure, and volume. In addition to original 25 hemodynamic outputs, we introduce 3 new outputs, including \textbf{systolic blood pressure, diastolic blood pressure, and mean arterial pressure}, bringing the total number of output variables to 28. For this work, we only predict a subset of output covariates highlighted in \textbf{Table 3}. A complete list of output are listed in \textbf{Table 3}.

\begin{itemize}
    \item \textbf{Systoblic Blood Pressure (SBP)} is defined as the highest measured arterial blood pressure while heart contracting. 
    \item \textbf{Diastolic Blood Pressure (DBP)} is defined as the lowest measured arterial blood pressure while heart contracting. 
    \item \textbf{Mean Arterial Pressure (MAP)} is defined as the average pressure in a patient’s arteries during one cardiac cycle as follow formula.
    $ MAP = \frac{2*DBP + 1* SBP}{3}$
\end{itemize}

\begin{table}[h!]
    \caption{Outputs of S-CVSim, $L_t$ denotes all the output of S-CVSim at time $t$; whereas, $L^{any}_t$ denotes a subset of output e.g. $L^{map}_t$ represents mean arterial pressure at time $t$. Covariates highlighted with \textbf{*} are the selected output. }
    \begin{center}
    \begin{tabular}{l|l}
    \textbf{OUTPUT COVARIATES}    & {}\\
    \hline \\
     \textbf{Left Ventricle Pressure*} & LVP \\
     \textbf{Left Ventricle Flow*} & LVQ\\
     Left Ventricle Volume & LVV \\
     \textbf{Left Ventricle Contractility*} & LVC \\
     \textbf{Right Ventricle Pressure*} & RVP \\
     \textbf{Right Ventricle Flow*} & RVQ \\
     Right Ventricle Volume & RVV \\
     \textbf{Right Ventricle Contractility*} & RVC\\
     \textbf{Central Venous Pressure*} & CVP\\
     Central Venous Flow & CVQ \\
     Central Venous Volume & CVV\\
     \textbf{Arterial Pressure*} & AP\\
     \textbf{Arterial Flow*} & AQ \\
     \textbf{Arterial Volume*} & AV\\
     Pulmonary Arterial Pressure & PAP\\
     Pulmonary Arterial Flow & PAQ\\
     Pulmonary Arterial Volume & PAV\\
     Pulmonary Venous Pressure & PVP\\
     Pulmonary Venous Flow & PVQ\\
     \textbf{Pulmonary Venous Volume*} & PVV\\     
     \textbf{Heart Rate*} & HR\\
     \textbf{Arteriolar Resistance*} & AR\\
     \textbf{Venous Tone*} & VT\\
     \textbf{Total Blood Volume*} & TBV\\
     \textbf{Intra-thoracic Pressure*} & PTH \\
     \textbf{Mean Arterial Pressure*} & MAP\\
     \textbf{Systolic Blood Pressure*} & SBP\\
     Diastolic Blood Pressure & DBP\\
    \end{tabular}
    \end{center}

    \label{tab:my_label}
\end{table}
\subsection{Simulation Process}
To obtain observational data for our intended purpose, we implemented and built two extra events, \textbf{Disease, $S_t$} and \textbf{Treatment, $A_t$}. Note that at each time $T$, both $S_t$ and $A_t$ could happen simultaneously but $S_t$ also happens before $A_t$. With these stochastic events, $A_t$ and $S_t$, $D_{obs}$ and $D_{CF}$ can be simulated and obtained on S-CVSim.

\subsubsection{Data Generation, $D_{o}, D_{c}$}
To simulate a patient trajectory under treatment strategy $g$, we obtain $D_{o}$ and $D_{c}$ with the following algorithm for each individual trajectory. $D_{o}$, training set, consists of 10k trajectories based on $g_{o}$. $D_{c}$, test set, consists of 1k trajectories based on $g_{c}$.
\begin{itemize}
\item Initialize input variables $V_1,...,V_N$ by drawing from independent uniform distributions using a predefined plausible physiological ranges for each variable as \textbf{Table 2} suggested.
\item For t in 0:K
\begin{itemize}
\item Generate $L_t^*$ as $F_{sim,t}(V,\bar{L}_{t-1}^*,A_{t-1})$, where $A_{-1}$ is taken to be $0$.
\item If $g$ = $g_{c}$ and t $\geq$ $\frac{K}{2}$:
\begin{itemize}
    \item Generate $A_t$ as $g_{c}(\bar{L}_t)$
\end{itemize}
\item Else
\begin{itemize}
    \item Generate $A_t$ as $g_{o}(\bar{L}_t)$
\end{itemize}
\end{itemize}
\end{itemize}

\subsubsection{ Disease Simulation,  $S_t$ }
We introduce the concept $S_t$ to simulate hemodynamic instability of cardiovascular system such as sepsis and bleeding at each timestep. $S_t$ consists of two events, \textbf{sepsis} and \textbf{blood loss}. In module of $S_t$, we denote that $P(S_t|L_t)=0.05$, and $P(Sepsis|S_t) = P(Blood Loss|S_t) = 0.5$; Note that blood loss and sepsis events are mutually exclusive in the simulation process at any time $t$. 
\begin{itemize}
    \item When sepsis happens, $L_{t+1}^{tpr} = \alpha_{tpr}*L_{t}^{tpr} \text{ where } 0 < \alpha_{tpr} \leq 0.7$, meaning $L_{t}^{tpr}$, total peripheral resistance, would decrease in $\alpha_{tpr}$ at next time $t+1$ 
    \item When blood loss happens, $L_{t+1}^{tbv} = \alpha_{tbv}*L_{t}^{tbv} \text{ where } 0 < \alpha_{tbv} \leq 0.95$, meaning that $L_{t}^{tbv}$, total blood volume would decrease in $\alpha_{tbv}$ at next time $t+1$.
\end{itemize}

\subsubsection{Treatment Simulation, $A_t$}
We developed two treatment strategy  $g_{o}$, policy for observational regime and $g_{c}$, policy for counterfactual policy, to simulate clinical treatment and validate our model. 
Under any given $g$, $A_t^{any}$ is defined as $g({L}_t)$. Similar to disease simulation, $A^{any}_t$ is either \textbf{ $A^{1}_t$, increasing quantity of total blood volume or $A^{2}_t$ increasing quantity of arterial resistance}. The probability of choosing fluids is equal to vasopressor but will not be administered at the same time. The dosage of $A_t$ depends on a subset of $L_t$ which indicates hemodynamic balance. More specific, since adequate blood pressure is an important clinical goal\cite{survivingSepsisCampaign}, we denote mean arterial pressure, MAP, of 65 mmHg and central venous pressure, CVP, of 10 mmHg as target goals. Therefore, We define $\Delta_{map,t}\equiv 65-L^{map}_t$ and $\Delta_{cvp,t}\equiv 10- L^{cvp}_t$ as proxies of how much dosage should be delivered. The following section will discuss the difference of $A_t$ between ${g_{CF}}$ and ${g_{obs}}$. 
\subsubsection{Observational Regime, ${g_{o}}$}
Under $g_{obs}$, probability and dosage of treatment are denoted as the followings
\begin{itemize}
    \item Probability of treatment, $P(A_t|L_t) = \frac{1}{1+e^{-x}} $ \text{ ,where } $x = C_1*\Delta_{map} + C_2*\Delta_{cvp} + C_0.$
    \item If we administer fluids, we generate the dose (in mL) $A_t^1 \sim max(0, \beta_1^1*\Delta_{map,t} + \beta_2^1*\Delta_{cvp,t} + \mathcal{N}(1500, 1000))$.
    \item If we administer vasopressors, we generate the dose $A_t^2 \sim max(0,\beta_1^2*\Delta_{map} + \beta_2^2*\Delta_{cvp} + \mathcal{N}(0,1))$ if $U\sim Uniform$. 
\end{itemize}
 
 \subsubsection{Counterfactual Regime, ${g_{c}}$}
Under $g_{CF}$, probability and dosage of treatment are denoted as the followings
 \begin{itemize}
     \item Probability of treatment, $P(A_t|L_t) = 1$ if and only if $L_{t}^{sbp} \leq 100$ and ShockIndex \cite{survivingSepsisCampaign}, $\frac{L_t^{hr}}{L_t^{sbp}}, \leq 0.8. $
     \item If we administer fluids, we generate the dose (in mL) $A_t^1 \sim max(0, \beta_1^1*\Delta_{map,t} + \beta_2^1*\Delta_{cvp,t})$.
     \item If we administer vasopressors, we generate the dose $A_t^2 \sim max(0,\beta_1^2*\Delta_{map} + \beta_2^2*\Delta_{cvp})$ if $U\sim Uniform$. 
 \end{itemize}
We experimented with multiple parameters and opted to use $ C0 = 0.02, C1 = 0.06, C2 = 0.24, \beta_1^1 = 10, \beta_2^1 = 60, \beta_1^2 = 0.1, \beta_2^2 = 0.15. $

\begin{figure}[h!]
    \centering
\includegraphics[width=0.31\textwidth,height=0.16\textheight]{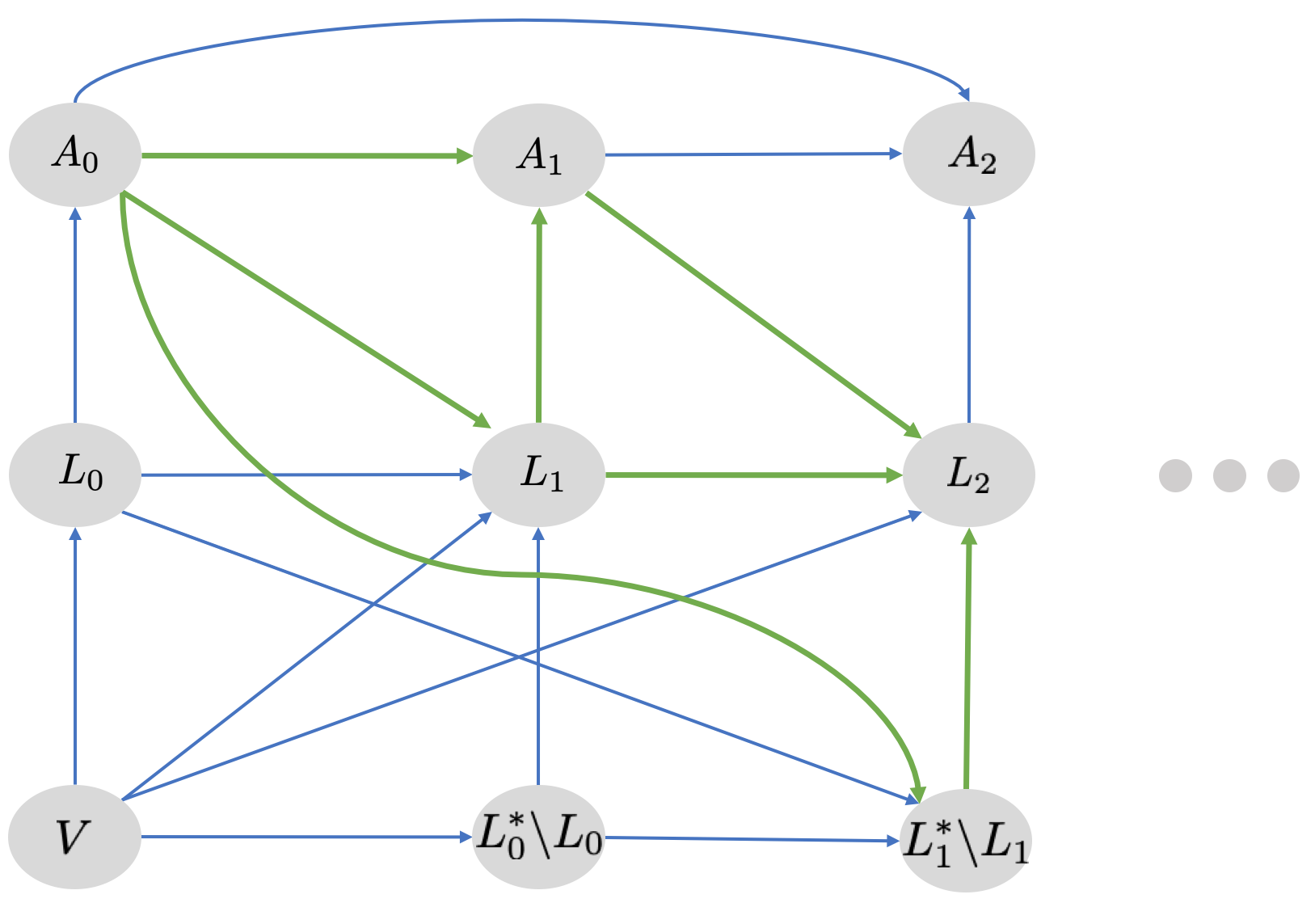}\\
    \caption{Causal DAG for $D_{o}$ generated under observational regime $g_{o}$. Green arrows denote directed paths from treatments to outcomes. Treatment only depends on current covariates. All covariates with arrows into treatment (i.e. MAP, CVP, and $\Delta_P$) are observed, satisfying the sequential exchangeability assumption as in Figure 1. }
    \label{dags}
\end{figure}
The DAG in Figure \ref{dags} depicts the causal structure of $D_{o}$. Note that only observed covariates have arrows pointing into treatment (treatment is actually only a function of CVP, MAP, and $\Delta_P$), so that there is no unobserved confounding and g-computation may be applied.

\end{document}